\documentclass{article}

\usepackage{PRIMEarxiv}

\usepackage[utf8]{inputenc} 
\usepackage[T1]{fontenc}    
\usepackage{hyperref}       
\usepackage{url}            
\usepackage{booktabs}       
\usepackage{amsfonts}       
\usepackage{nicefrac}       
\usepackage{microtype}      
\usepackage{lipsum}
\usepackage{fancyhdr}       
\usepackage{graphicx}       
\graphicspath{{media/}}     

\usepackage{natbib}
\usepackage{xspace}
\usepackage{multirow}
\usepackage{todonotes}
\usepackage{makecell}

\newcommand{\HyperGLUEpt}{PORTULAN ExtraGLUE\xspace}

\newcommand{\ptpt}{pt-PT\xspace}
\newcommand{\ptbr}{pt-BR\xspace}

\title{\HyperGLUEpt\ Datasets and Models:\\Kick-starting a Benchmark for the Neural Processing of Portuguese}

\author{
  Tomás Freitas Osório$^{\dagger}$, Bernardo Leite$^{\dagger}$, Henrique Lopes Cardoso$^{\dagger}$, \\
  \textbf{Luís Gomes}$^{\ddagger}$, \textbf{João Rodrigues}$^{\ddagger}$, \textbf{Rodrigo Santos}$^{\ddagger}$, \textbf{António Branco}$^{\ddagger}$ \vspace{5pt} \\
  $^{\dagger}$Laboratório de Inteligência Artificial e Ciência de Computadores (LIACC)\\
  Faculdade de Engenharia da Universidade do Porto\\
  Rua Doutor Roberto Frias, s/n, Porto, Portugal\\
  \texttt{tomas.s.osorio@gmail.com, \{bernardo.leite, hlc\}@fe.up.pt} \vspace{5pt} \\
  $^{\ddagger}$University of Lisbon\\
  NLX---Natural Language and Speech Group, Dept of Informatics\\
  Faculdade de Ciências (FCUL), Campo Grande, 1749-016 Lisboa, Portugal\\
  \texttt{\{lmdgomes, jarodrigues, rsdsantos, antonio.branco\}@fc.ul.pt} \\
}

\begin{document}
\maketitle

\begin{abstract}
Leveraging research on the neural modelling of Portuguese, we contribute a collection of datasets for an array of language processing tasks and a corresponding collection of fine-tuned neural language models on these downstream tasks.
To align with mainstream benchmarks in the literature, originally developed in English, and to kick start their Portuguese counterparts, the datasets were machine-translated from English with a state-of-the-art translation engine. The resulting \HyperGLUEpt benchmark is a basis for research on Portuguese whose improvement can be pursued in future work.
Similarly, the respective fine-tuned neural language models, developed with a low-rank adaptation approach, are made available as baselines that can stimulate future work on the neural processing of Portuguese.
All datasets and models have been developed and are made available for two variants of Portuguese: European and Brazilian.
\end{abstract}


\section{Introduction}

Neural language models are pervasive in Natural Language Processing (NLP) applications and have radically changed the state-of-the-art since the Transformer architecture \citep{NIPS2017_3f5ee243} was proposed. This has given rise to encoder \citep{devlin-etal-2019-bert}, decoder \citep{radford2018improving}, and encoder-decoder architectures \citep{2020t5}.
To support the development of such models, several benchmarks have been created to assess their performance in several downstream tasks \citep{wang-etal-2018-glue,Wang2019-superglue}. However, most research in NLP has focused on the English language \citep{10.33011/lilt.v6i.1239}, and as a consequence, many other languages lack sufficient resources -- in particular benchmarks for neural language models.

Developing benchmark datasets is hard, usually demanding labelling by experts, especially for complex semantic-level tasks.
An alternative path that has been resorted to in the literature is to rely on state-of-the-art Machine Translation (MT) to produce dependable datasets, namely those that support the evaluation of neural models in downstream tasks \citep{conneau-etal-2018-xnli,eger-etal-2018-cross,yang-etal-2019-paws,carrino-etal-2020-automatic,dhoffschmidt-etal-2020-fquad,shavrina-etal-2020-russiansuperglue,carvalho_2021_qa_pt,sousa2021argmining,zagar-robnik-sikonja-2022-slovene}. Though possibly imperfect, such datasets can fit the purpose of greatly leveraging research in less-resourced languages, possibly complemented with human-curated test sets.

In this paper, we contribute to enriching the set of benchmarks publicly available for Portuguese by relying on MT applied to tasks from the well-known GLUE \citep{wang-etal-2018-glue} and SuperGLUE \citep{Wang2019-superglue} benchmarks, which were originally developed for English. 
We discuss the issues encountered with our approach and provide versions of several tasks for European (\ptpt) and Brazilian (\ptbr) Portuguese, which altogether we named \HyperGLUEpt.

As a way of their practical validation, for most tasks, we include experimental evaluation of different Portuguese language models fine-tuned with the respective datasets. Hence, for many of them, these will be the first models to address that task in Portuguese, and we thus contribute the first baselines for them.
To that end, we resort to the encoder Albertina language model \citep{rodrigues2023advancing} and the low-rank adaptation approach \citep{hu2022lora}. The resulting fine-tuned language models for these tasks are openly distributed as open source under an open license.

\section{Related Work}

Producing benchmarks to evaluate language models in downstream tasks is a daunting endeavour. The more complex the task, the more difficult it is to produce quality data that can be used to train models in a fine-tuning approach and test their capabilities. While highly resourced languages, such as English, include quite elaborate benchmarks \citep{wang-etal-2018-glue,Wang2019-superglue}, few evaluation datasets are available for other, less-resourced languages.\footnote{For instance, treebank annotations \citep{nivre-etal-2020-universal} are available, but do not comprise benchmarks \textit{per se}.}
The particular case of Portuguese is a paradigmatic example, with only a few tasks being available for this purpose \citep{fonseca2016assin,real2020assin,santos-etal-2006-harem,freitas-etal-2010-second}.

A few examples of manually produced multilingual parallel corpora exist \citep{yang-etal-2019-paws,artetxe-etal-2020-cross,ponti-etal-2020-xcopa,sen-etal-2022-mintaka}, as well as collections of tasks in multiple languages \citep{srivastava2023beyond}.
At the same time, machine translation has come to a point in which it can be useful to create corpora that, while lacking human curation, can, up to a certain extent, be used to evaluate language models in the target languages \citep{conneau-etal-2018-xnli,eger-etal-2018-cross,yang-etal-2019-paws,carrino-etal-2020-automatic,dhoffschmidt-etal-2020-fquad}. Some have been created to allow cross-lingual evaluation of pre-trained encoders \citep{pmlr-v119-hu20b,liang-etal-2020-xglue}.

State-of-the-art MT systems still struggle to produce accurate translations in several situations.
Short texts, for instance, often lack enough context to obtain proper translations \citep{wan-etal-2022-challenges}. Because of this, translation at the sentence level often falls short of translating longer texts, which provide more context \citep{jin-etal-2023-challenges}.
Translating from mostly gender-poor to gender-rich languages is also often a source of translation errors \citep{savoldi-etal-2021-gender}.
Idioms are among the most intricate artefacts for MT systems, which tend to over-generate compositional and literal translations \citep{dankers-etal-2022-transformer}.
Additionally, translation-based data can arguably be seen as a dialect of the target language \citep{Volansky2013translationese,artetxe-etal-2020-translation}, with the possible effect of over-estimating the performance in the target language of models trained on such data. 
Still, MT has progressed notably over the last few years; it can, we believe, be used to produce datasets that are useful as a proxy in assessing the comparative merits of different (monolingual) language models.

Following this trend, some works have focused on leveraging MT to produce corpora in Portuguese \citep{carvalho_2021_qa_pt,sousa2021argmining}.
In this paper, we leverage state-of-the-art MT in producing Portuguese variants of several GLUE \citep{wang-etal-2018-glue} and SuperGLUE \citep{Wang2019-superglue} tasks.
Similar efforts have been made for other languages \citep{shavrina-etal-2020-russiansuperglue,zagar-robnik-sikonja-2022-slovene}.

In tandem with developing and making these datasets available, and as a way of their practical validation, we also release low-ranked adaptations \citep{hu2022lora} of Albertina-based models \citep{rodrigues2023advancing}, arguably the best open encoder models for both European and Brazilian Portuguese available at the time of this writing.

Low-ranked adaptations (LoRA) reduce the number of training parameters, alleviating storage requirements for language models adapted to specific tasks while outperforming other fine-tuning techniques. For that, the weights of the pre-trained model are frozen, and two additional weight matrices are used to adapt the model to the downstream task. After training, such weights can be merged with the frozen weights so that no latency is added at inference time, which is a main advantage compared to other low-rank adapters \citep{pmlr-v97-houlsby19a,mahabadi2021compacter,he2022towards}. Concerning LoRA, more recent proposals \citep{valipour-etal-2023-dylora,audibert-etal-2023-low} rely on the GLUE benchmark \citep{wang-etal-2018-glue} to report improvements.

\section{General Language Understanding Evaluation Benchmarks}

The General Language Understanding Evaluation (GLUE) tasks are meant to measure the progress toward general-purpose language understanding technologies for English. Both GLUE and SuperGLUE are aggregations of existing public datasets accompanied by a single-number performance metric and an analysis toolkit.
The tasks included in these benchmarks can be grouped as follows\footnote{We superscript each task regarding its inclusion in (G)LUE, (S)uperGLUE, or both.}.

\subsection{Single sentence tasks}

The Corpus of Linguistic Acceptability (\textbf{CoLA})\textsuperscript{G} \citep{warstadt-etal-2019-neural} is a task including sentences annotated for grammatical acceptability by experts in linguistics.
The Stanford Sentiment Treebank (\textbf{SST-2})\textsuperscript{G} \citep{socher-etal-2013-recursive}, in turn, is a task for predicting the sentiment polarity of movie reviews.

\subsection{Similarity tasks}

The Microsoft Research Paraphrase Corpus (\textbf{MRPC})\textsuperscript{G} \citep{dolan-brockett-2005-automatically} is a task for determining whether a pair of sentences are mutual paraphrases.
Quora Question Pairs (\textbf{QQP})\textsuperscript{G,}\footnote{\url{https://quoradata.quora.com/First-Quora-Dataset-Release-Question-Pairs}} is a task for determining whether a pair of questions are semantically equivalent.
The Semantic Textual Similarity Benchmark (\textbf{STS-B})\textsuperscript{G} \citep{cer-etal-2017-semeval} is a task for predicting a similarity score (from 1 to 5) for each sentence pair.
Word-in-Context (\textbf{WiC})\textsuperscript{S} \citep{pilehvar-camacho-collados-2019-wic} comprises a word sense disambiguation task, where given two sentences containing a polysemous target word, the aim is to determine whether the word is used in the same sense in both sentences.

\subsection{Inference tasks}

The Multi-Genre Natural Language Inference Corpus (\textbf{MNLI})\textsuperscript{G} \citep{williams-etal-2018-broad} is a task to determine if a given premise sentence entails, contradicts, or is neutral to a hypothesis sentence; the task includes matched (in-domain) and mismatched (cross-domain) validation and test sets.
Question NLI (\textbf{QNLI})\textsuperscript{G} \citep{rajpurkar-etal-2016-squad} is a question-answering task converted to determine whether the context sentence contains the answer to the question.
Recognizing Textual Entailment (\textbf{RTE})\textsuperscript{GS} is a task for determining whether a premise sentence entails a hypothesis sentence.
Winograd Natural Language Inference (\textbf{WNLI})\textsuperscript{G} \citep{Levesque2012-wsc} is a pronoun resolution task formulated as sentence pair entailment classification where, in the second sentence, the pronoun is replaced by a possible referent.
Similarly, the Winograd Schema Challenge (\textbf{WSC})\textsuperscript{S}
is a co-reference resolution task also formulated as sentence pair entailment classification, where each example comprises a sentence and a pair pronoun-noun, the objective being to determine if they are co-referent.
CommitmentBank (\textbf{CB})\textsuperscript{S} \citep{de-Marneffe_Simons_Tonhauser_2019-CB} comprises short texts with embedded clauses; one such clause is extracted as a hypothesis and should be classified as neutral, entailment or contradiction.

GLUE and SuperGLUE also include an expert-constructed diagnostic dataset covering different types of linguistic phenomena.
Broadcoverage Diagnostics (\textbf{AX$_b$})\textsuperscript{GS} \citep{wang-etal-2018-glue} is a Natural Language Inference (NLI) task designed to test models across a wide spectrum of linguistic, commonsense, and world knowledge; each instance contains a sentence pair labeled with entailment or not entailment.
Winogender Schema Diagnostics (\textbf{AX$_g$})\textsuperscript{S} \citep{rudinger-etal-2018-gender} is a similar task, designed to measure gender bias, where each premise sentence includes a male or female pronoun and a hypothesis includes a possible referent for the pronoun.

\subsection{Question-answering tasks}

Boolean Questions (\textbf{BoolQ})\textsuperscript{S} \citep{clark-etal-2019-boolq} is a question-answering task where yes/no questions are given for short text passages.
In the Multi-Sentence Reading Comprehension (\textbf{MultiRC})\textsuperscript{S} task \citep{khashabi-etal-2018-looking}, given a context paragraph, a question, and an answer, the goal is to determine whether the answer is true; for the same context and question, more than one answer may be correct.
In the Reading Comprehension with Commonsense Reasoning Dataset (\textbf{ReCoRD})\textsuperscript{S}, each sample is a multiple-choice question including a news article passage and a Cloze-style question with one entity masked out; the aim is to predict the masked entity from a list of alternatives.

\subsection{Reasoning tasks}

Choice of Plausible Alternatives (\textbf{COPA})\textsuperscript{S} \citep{gordon-etal-2012-semeval} is a casual reasoning task: given a premise, two choices, and a cause/effect prompt, the system must choose one of the choices.

\section{\HyperGLUEpt}

Creating a Portuguese version of the tasks introduced in the previous section via machine translation (MT) requires a thoughtful understanding of the nature of each task, together with the limitations of the selected MT engine.
While we are aware that, for a small subset of these tasks, Portuguese-translated versions have already been created \citep{rodrigues2023advancing}, such considerations have not been taken into account.
In fact, 
the inner workings of MT and the differences between languages (in our case, English and Portuguese) may impact the validity of the gold labels in supervised tasks. This is something we analyze in this section before providing details on the \HyperGLUEpt\ datasets we obtained.

For MT, we use DeepL\footnote{All the examples in this section are obtained via DeepL's web interface (\url{https://www.deepl.com/translator}) at the time of writing.}, a commercial MT tool that tailors translation to two Portuguese variants, European (\ptpt) and Brazilian (\ptbr).

\subsection{More than translation}\label{sec:more-than-translation}

Both statistical and neural sequence-to-sequence MT models are affected by language model probabilities. As a side effect, ill-formed or ungrammatical source sentences are affected in the translation process, hindering the faithfulness of the output in the target language as a direct counterpart of the input in the source language. MT has been used in grammatical error correction \citep{rozovskaya-roth-2016-grammatical,kementchedjhieva-sogaard-2023-grammatical}.

For this reason, we abstain from machine-translating the CoLA dataset, as the obtained translation may easily corrupt the target labels.
As an example, the sentence ``They drank the pub'' (linguistically $ungrammatical$) is translated to \ptbr\ as ``Eles beberam \textit{no} bar'' (``They drank \textit{in the} pub'', $grammatical$).
As another example, the sentence ``The professor talked us'' ($ungrammatical$) is translated to \ptpt\ as ``O professor falou\textit{-nos}'' (``The professor talked \textit{to} us'', $grammatical$).

\subsection{Gendered nouns and pronoun resolution}\label{sec:gendered-nouns-and-pronoun-resolution}

English common nouns do not express grammatical gender. On the other hand, Portuguese common nouns do and are used with corresponding gendered determiners (as opposed to English gender-neutral \textit{the} or \textit{a}).
This exacerbates the difficulty of properly addressing pronoun reference resolution, given that third-person singular pronouns (and also plural in Portuguese) are gendered.
Tasks specifically dealing with pronoun resolution or evaluating the gender robustness of language models are thus prone to corruption via MT.
These include WNLI, WSC, and AX$_g$. While we provide translated versions of WNLI and AX$_g$, we conduct error analysis to diagnose the quality level of their Portuguese versions.

An example of a translation issue in WNLI is as follows: ``Tom said "Check" to Ralph as he took his bishop'' / ``Tom said "Check" to Ralph as he took Ralph's bishop'' is translated to \ptpt\ as ``O Tomás disse "Xeque" ao Rafa quando \textit{este} lhe tirou o bispo'' / ``O Tomás disse "Xeque" ao Rafa quando tirou o bispo ao Rafa''. The first sentence in the pair is wrongly translated (\textit{este} means \textit{the latter}), and even though it does not make sense, the target label should change from $entailment$ to $not\_entailment$.

For WSC, the situation is more critical, as parts of the input are isolated words (usually nouns and pronouns). Thus, obtaining a proper Portuguese equivalent requires more than MT. An example is as follows: from ``The \textit{mothers} of Arthur and Celeste have come to the town to fetch them. They are very happy to have them back, but they scold \textit{them} just the same because they ran away'', we want to determine whether the italicised words are co-referent. In this example, there is no separated word matching \textit{them} (which should translate to \textit{eles}) in the translation ``As mães do Artur e da Celeste vêm buscá-los à cidade. Estão muito contentes por \textit{os} terem de volta, mas repreendem\textit{-nos} na mesma por terem fugido''.

AX$_g$ focuses on gender bias, explicitly combining both concerns expressed above.
For instance, the hypothesis ``The investigator tried to get in contact'' is translated into Portuguese as ``O investigador tentou entrar em contacto''; its possibly accompanying premises ``The investigator wanted to interview the witness in person, but [he | she] was unable to get in contact'' are translated into Portuguese as ``O investigador queria entrevistar a testemunha pessoalmente, mas não conseguiu entrar em contacto com \textit{ela}'' (for he), or to ``O investigador queria entrevistar a testemunha pessoalmente, mas \textit{ela} não conseguiu entrar em contacto'' (for \textit{she}). In the latter case, limiting the possible referents of pronoun \textit{ela} (she) -- the only feminine noun is \textit{testemunha} (witness), since \textit{investigador} (investigator) is masculine in Portuguese -- renders the $entailment$ label wrong, as it should be changed to $not\_entailment$.

\subsection{Named entities}\label{sec:named-entities}

Another issue we have encountered when using DeepL is the non-deterministic translation of common or proper names, which might make fine-tuning models in these datasets harder or even impact label quality.
Consider the following example, taken from WNLI:
``\textit{Jane} gave \textit{Joan} candy because she wasn't hungry'' / ``\textit{Jane} wasn't hungry'' is translated to \ptpt\ as ``A \textit{Joana} deu doces à \textit{Joana} porque ela não tinha fome'' / ``A \textit{Joana} não tinha fome''; in this example, one of the distinct proper names is lost.
The reverse can also happen: ``Bill passed the half-empty plate to \textit{John} because he was full'' / ``\textit{John} was full'' is translated to \ptpt\ as ``O Bill passou o prato meio vazio ao \textit{John} porque estava cheio'' / ``O \textit{João} estava cheio''; in this case, a single entity, \textit{John}, is either kept or translated to \textit{João} in the same short text.

As another example from the same dataset, now concerning the same common noun being translated differently, ``I couldn't put the \textit{pot} on the shelf because it was too tall'' / ``The \textit{pot} was too tall''. is translated to \ptpt\ as ``Não podia colocar a \textit{panela} na prateleira porque era demasiado alta'' / ``O \textit{pote} era demasiado alto''.

These issues may be prevalent in every dataset, particularly in \ptpt\ variants.

\subsection{Machine-translated tasks}

The set of datasets that have been translated and are part of \HyperGLUEpt\footnote{Made available at \url{https://huggingface.co/datasets/PORTULAN/extraglue}.} are included in Table~\ref{tab:pt-datasets}.
As mentioned in Sections~\ref{sec:more-than-translation} and~\ref{sec:gendered-nouns-and-pronoun-resolution}, we leave out the CoLA and WSC datasets.

\begin{table*}[t]
    \centering
    \begin{tabular}{lccccccccc}
        \hline
        \textbf{Task} & \textbf{Train} & \textbf{Val} & \textbf{Test} & \textbf{Tokens (en)} & \textbf{Version} & \textbf{Tokens (pt)} & \textbf{mt$_e$} & \textbf{lab$_e$} & \textbf{low$_q$} \\
        \hline
        \hline
        \multirow{2}{*}{SST-2} & \multirow{2}{*}{67.3k} & \multirow{2}{*}{872} & \multirow{2}{*}{1.82k} & \multirow{2}{*}{686.1k} & \ptpt & 725.3k & 4\% & 0\% & 0\% \\
             &  &  &  &  & \ptbr & 724.9k & 4\% & 0\% & 0\% \\
        \hline
        \hline
        \multirow{2}{*}{MRPC} & \multirow{2}{*}{3.67k} & \multirow{2}{*}{408} & \multirow{2}{*}{1.73k} & \multirow{2}{*}{254.3k} & \ptpt & 287.2k & 4\% & 0\% & 2\% \\
             &  &  &  &  & \ptbr & 284.7k & 6\% & 0\% & 2\% \\
        \hline
        \multirow{2}{*}{STS-B} & \multirow{2}{*}{5.75k} & \multirow{2}{*}{1.5k} & \multirow{2}{*}{1.38k} & \multirow{2}{*}{197.5k} & \ptpt & 220.6k & 2\% & 0\% & 0\% \\
             &  &  &  &  & \ptbr & 217.8k & 2\% & 0\% & 0\% \\
        \hline
        \hline
        \multirow{2}{*}{\makecell[l]{MNLI\\\_matched}} & \multirow{2}{*}{--} & \multirow{2}{*}{9.82k} & \multirow{2}{*}{9.8k} & \multirow{2}{*}{649.4k} & \ptpt & 660.6k & 0\% & 0\% & 0\% \\
             &  &  &  &  & \ptbr & 661.4k & 4\% & 0\% & 0\% \\
        \hline
        \multirow{2}{*}{\makecell[l]{MNLI\\\_mismatched}} & \multirow{2}{*}{--} & \multirow{2}{*}{9.83k} & \multirow{2}{*}{9.85k} & \multirow{2}{*}{680.6k} & \ptpt & 710.3k & 6\% & 0\% & 0\% \\
             &  &  &  &  & \ptbr & 705.3k & 4\% & 0\% & 0\% \\
        \hline
        \multirow{2}{*}{QNLI} & \multirow{2}{*}{105k} & \multirow{2}{*}{5.46k} & \multirow{2}{*}{5.46k} & \multirow{2}{*}{4.82M} & \ptpt & 5.22M & 2\% & 2\% & 2\% \\
             &  &  &  &  & \ptbr & 5.14M & 0\% & 0\% & 0\% \\
        \hline
        \multirow{2}{*}{RTE} & \multirow{2}{*}{2.49k} & \multirow{2}{*}{277} & \multirow{2}{*}{3k} & \multirow{2}{*}{333.8k} & \ptpt & 364.4k & 2\% & 0\% & 0\% \\
             &  &  &  &  & \ptbr & 360.8k & 2\% & 0\% & 0\% \\
        \hline
        \multirow{2}{*}{WNLI} & \multirow{2}{*}{635} & \multirow{2}{*}{71} & \multirow{2}{*}{146} & \multirow{2}{*}{29.7k} & \ptpt & 30.2k & 6\% & 4\% & 4\% \\
             &  &  &  &  & \ptbr & 29.5k & 8\% & 6\% & 6\% \\
        \hline
        \multirow{2}{*}{CB} & \multirow{2}{*}{250} & \multirow{2}{*}{56} & \multirow{2}{*}{250} & \multirow{2}{*}{43.3k} & \ptpt & 40.4k & 6\% & 2\% & 2\% \\
             &  &  &  &  & \ptbr & 40.5k & 8\% & 2\% & 4\% \\
        \hline
        \multirow{2}{*}{AX$_b$} & \multirow{2}{*}{--} & \multirow{2}{*}{--} & \multirow{2}{*}{1.1k} & \multirow{2}{*}{40.2k} & \ptpt & 43.3k  & 20\% & 4\% & 14\% \\
             &  &  &  &  & \ptbr & 42.7k & 20\% & 4\% & 12\% \\
        \hline
        \multirow{2}{*}{AX$_g$} & \multirow{2}{*}{--} & \multirow{2}{*}{--} & \multirow{2}{*}{356} & \multirow{2}{*}{8.7k} & \ptpt & 8.9k & 22\% & 10\% & 10\% \\
             &  &  &  &  & \ptbr & 8.8k & 20\% & 6\% & 8\% \\
        \hline
        \hline
        \multirow{2}{*}{BoolQ} & \multirow{2}{*}{9.43k} & \multirow{2}{*}{3.27k} & \multirow{2}{*}{3.25k} & \multirow{2}{*}{1.93M} & \ptpt & 2.07M & 22\% & 2\% & 12\% \\
             &  &  &  &  & \ptbr & 2.06M & 18\% & 2\% & 8\% \\
        \hline
        \multirow{2}{*}{MultiRC} & \multirow{2}{*}{27.2k} & \multirow{2}{*}{4.85k} & \multirow{2}{*}{9.69k} & \multirow{2}{*}{12.99M} & \ptpt & 13.69M & 10\% & 2\% & 2\% \\
             &  &  &  &  & \ptbr & 13.65M & 10\% & 4\% & 4\% \\
        \hline
        \hline
        \multirow{2}{*}{CoPA} & \multirow{2}{*}{400} & \multirow{2}{*}{100} & \multirow{2}{*}{500} & \multirow{2}{*}{19.5k} & \ptpt & 18.6k  & 2\% & 2\% & 2\% \\
             &  &  &  &  & \ptbr & 19.3k & 2\% & 2\% & 2\% \\
        \hline
    \end{tabular}
    \caption{\HyperGLUEpt datasets. For each task, we include the size of each partition, the number of tokens in each Portuguese variant, and results from the sample analysis in percentages (mt$_e$ = machine translation errors, lab$_e$ = corrupted labels, and low$_q$ = low-quality translated samples).}
    \label{tab:pt-datasets}
\end{table*}

For MNLI, we provide translations only for the matched and mismatched validation and test sets due to the excessive size of the training set\footnote{The training set for MNLI contains 393k rows.}.
For the same reason, we do not translate the QQP dataset\footnote{QQP includes a total of 795k rows.}.

Given the nature of the WiC task (based on word sense disambiguation), we posit that a (human or machine) translated version of this dataset is not viable and thus leave it out.
Finally, given the focus of the ReCoRD task on named entities and the issues encountered and described in Section~\ref{sec:named-entities}, we abstain from translating this dataset as well.

To improve translation quality, we concatenate each dataset entry's textual columns with a line break. This ensures that the MT model can access as much context as is available (which may be critical for datasets with very short text spans) and is in line with previous findings \citep{artetxe-etal-2020-translation}.

As it can be seen in Table~\ref{tab:pt-datasets}, the number of tokens varies among the Portuguese language variants. To better assess how different these are in the resulting machine-translated datasets, we calculate the BLEU score \citep{papineni-etal-2002-bleu} between both variants. For that, we rely on 4-grams; BLEU is calculated independently for each feature (text column in a dataset) and then averaged for the whole dataset. The BLEU score averaged over both directions (\ptpt $\rightarrow$ \ptbr and \ptbr $\rightarrow$ \ptpt) and for all datasets is 57.3, with the lowest value of 46.7 on the CoPA dataset and the highest of 64.5 on RTE. 
These values demonstrate that there are significant differences between the translations obtained for each variant via DeepL.

To assess the quality of each machine-translated dataset, we resort to sampling $50$ randomly selected examples, which were manually checked by three of the authors\footnote{Portuguese native speakers and fluent in English.} for translation correctness and target label consistency.
The rightmost columns in Table~\ref{tab:pt-datasets} show the results of this analysis: obvious translation errors, label corruption, and low-quality entries that should be removed from the dataset, given its nature.

The main translation problems we have observed concern pronoun resolution or gender issues (as already emphasized in Section~\ref{sec:gendered-nouns-and-pronoun-resolution}), idiomatic expressions, inconsistent translations in pairs of sentences, and a few cases of `hallucinations,'
among other problematic mistranslations.
In some cases, these problems have an impact on the correctness of the labels (mainly in WNLI and AX$_g$); in other cases, they mostly imply a dataset of lesser quality (such as in AX$_b$ and BoolQ).
In the specific case of AX$_g$, even when the translation is correct, it does not do justice to the nature of the task, which loses its purpose (e.g., \textit{his}/\textit{her} translate the same way to Portuguese).

Despite these problems, machine translation errors amount to only an average of 8\%, with a mode as low as 2\%. Label errors are even lower, with an average of 2\% and a zero mode.
We did not observe relevant differences between Portuguese language variants.

\section{Albertina LoRA Models}

We train and make available a set of fine-tuned low-rank adaptations of Albertina-based language models\footnote{Made available at [\textit{url temporarily removed to preserve anonymity during review}]}.
For several \HyperGLUEpt datasets, we fine-tune a 1.5B Albertina language model for two Portuguese variants, European (\ptpt) and Brazilian (\ptbr).
The resulting models are a practical validation for the created datasets.

\subsection{Set up}

First, we adapt each task example for tokenization regarding their input components. For this, we concatenate the input features with a special token separator. On the MRPC and STS-B similarity tasks, we concatenate the first and second sentences. On the CB and RTE inference tasks, the hypothesis and premise; on QNLI, the sentence and question. For the BoolQ Question-answering task, we concatenate the passage and question; for MultiRC, the paragraph, question, and answer, truncating the paragraph if needed. For the CoPA reasoning task, we concatenate the premise and question and then join with each choice, resulting in two inputs. During tokenization, we truncate the examples with a maximum context length of 128 tokens, except in MultiRC, which uses 256 tokens.

\begin{table}[t]
    \centering
    \begin{tabular}{cc}
        \hline
        \textbf{Hyper-parameter} & \textbf{Value} \\
        \hline
        \hline
        r & $8$\\
        \hline
        alpha & $32$\\
        \hline
        dropout & $0.05$\\
        \hline
        batch size & $8$\\
        \hline
        learning rate & $2 \times 10^{-5}$\\
        \hline
        weight decay & $0.05$\\
        \hline
    \end{tabular}
    \caption{LoRA hyper-parameters.}
    \label{tab:LoRA-hyperparameters}
\end{table}

After tokenization, we apply a low-rank adapter \citep{hu2022lora} with the hyper-parameters shown in Table~\ref{tab:LoRA-hyperparameters}.
Due to hardware limitations, it was unfeasible to perform a grid search on these hyper-parameters.
We chose the current hyper-parameters by resorting to small-scale exploratory experiments.
Because several datasets lack test labels, we fine-tuned models on the training split and evaluated them on the validation split.

\subsection{Results}

\begin{table*}[!ht]
    \centering
    \begin{tabular}{lcc|cc|c}
        \hline
        \multirow{2}{*}{\textbf{Task}} & \multicolumn{2}{c|}{\textbf{Albertina 1.5B}} & \multicolumn{2}{c|}{\textbf{XLM-RoBERTa-XL}} & \textbf{DeBERTa-V2-XXLarge} \\
        & \textbf{\ptpt} & \textbf{\ptbr} & \textbf{\ptpt} & \textbf{\ptbr} & \textbf{en} \\
        \hline
        \hline
        Single sentence & & & & & \\
        \cline{1-1}
        SST-2 & 0.9392 & 0.9450 & 0.9323 & 0.9392 & 0.9633 \\
        \hline
        \hline
        Similarity & & & & & \\
        \cline{1-1}
        MRPC & 0.8969 & 0.9184 & 0.8696 & 0.8651 & 0.9266 \\
        STS-B & 0.8905 & 0.8940 & 0.8743 & 0.8734 & 0.9170 \\
        \hline
        \hline
        Inference & & & & & \\
        \cline{1-1}
        QNLI & 0.9398 & 0.9361 & 0.9237 & 0.9237 & 0.9608 \\
        RTE & 0.7870 & 0.7978 & 0.6571 & 0.6606 & 0.8917 \\
        WNLI & 0.6197 & 0.6901 & 0.5634 & 0.5634 & 0.7887 \\
        CB & 0.8385 & 0.8554 & 0.6280 & 0.6160 & 89.36 \\
        \hline
        \hline
        QA & & & & & \\
        \cline{1-1}
        BoolQ & 0.7456 & 0.7807 & 0.6538 & 0.6587 & 0.8900 \\
        MultiRC & 0.7257 & 0.7169 & 0.6926 & 0.6925 & 0.8243 \\
        \hline
        \hline
        Reasoning & & & & & \\
        \cline{1-1}
        CoPA & 0.8500 & 0.8200 & 0.5000 & 0.5600 & 0.9200 \\
        \hline
    \end{tabular}
    \caption{Evaluation scores on validation sets for both variants regarding the different categories of datasets (Single Sentence, Similarity, Inference, Question-Answering, and Reasoning). Performance on SST-2, QNLI, RTE, WNLI, BoolQ, and CoPA is measured with accuracy; on MRPC, CB, and MultiRC with F1; and on STS-B with Pearson. 
    For comparison, we include results for the multilingual XLM-RoBERTa-XL 3.5B model, fine-tuned using the same LoRA approach.
    For reference, we also include results for English by applying LoRA to the DeBERTa-V2-XXLarge 1.5B model (based on which Albertina has been developed).}
    \label{tab:results}
\end{table*}

The fine-tuning results are presented in Table~\ref{tab:results}.
All these models are the first baselines for the tasks regarding these new datasets.

Comparing the empirical results between the two variants (\ptpt and \ptbr), we observe that the \ptbr variant achieves better scores than the \ptpt variant in seven tasks (SST-2, MRPC, STS-B, RTE, WNLI, CB, and BoolQ), while the \ptpt variant has better scores in three tasks (QNLI, MultiRC, and CoPA).
It is worth noting, however, that the differences are marginal in most cases. The larger discrepancies are observed for the WNLI, BoolQ and CoPA tasks. The first two tasks yield better results with the \ptbr variant, whereas the CoPA task achieves a better outcome in the \ptpt variant.

We can also compare the results with those available for a subset of tasks and the current state-of-the-art Albertina models, as reported in \citet{rodrigues2023advancing}. For the \ptpt variant: 
in MRPC we obtain 0.8969 accuracy compared to 0.9171 in the original 900M Albertina model;
in STS-B we obtain a Pearson correlation of 0.8905 compared to Albertina's 0.8801;
in RTE we obtain 0.7870 accuracy against .8339;
and in WNLI we obtain 0.6197 accuracy against 0.4225.
For the \ptbr variant:
in MRPC we obtain 0.9184 accuracy compared to 0.9071 in the original 900M Albertina model; in STS-B we obtain a Pearson correlation of 0.8940 compared to Albertina's 0.8910;
in RTE we obtain 0.7978 accuracy against 0.7545;
and in WNLI we obtain 0.6901 accuracy against 0.4601.
We note, however, that the translations of these tasks in \HyperGLUEpt may differ from the translations used by the authors of the Albertina model for their evaluations. This is certainly true for the \ptbr variant, as the MT model used differed.

Table~\ref{tab:results} also includes the results obtained by fine-tuning the multilingual XLM-RoBERTa-XL\footnote{\url{https://huggingface.co/facebook/xlm-roberta-xl}} model \citep{conneau-etal-2020-unsupervised} following the same LoRA approach. XLM-RoBERTa-XL is significantly larger (3.5B parameters) than Albertina 1.5B.
Even so, we note the benefits of using monolingual models when comparing such results with our Albertina 1.5B LoRA models. In fact, we observe improvements in Albertina 1.5B LoRA models for all tasks and in both Portuguese variants. In some cases, improvements are significant.

When comparing with the DeBERTa\footnote{\url{https://huggingface.co/microsoft/deberta-v2-xxlarge}} \citep{He:2021:DeBERTa} model (the foundation model for Albertina) applied to the original English datasets, the results of our low-rank adapters on the \HyperGLUEpt datasets fall behind in most cases. This is expected for at least two reasons: first, Albertina was pre-trained with far fewer data than DeBERTa; second, we rely on machine translation to obtain the datasets for the tasks, which, as discussed before, isn't without issues.
Tasks exhibiting significant differences in performance include WNLI, which, as explained in Section~\ref{sec:gendered-nouns-and-pronoun-resolution}, has issues related to pronoun resolution.

\section{Conclusion}

We contribute an open benchmark suite to support the development of the neural processing of Portuguese. In this initial version, this suite comprises 14 datasets for downstream tasks of various types, including single sentence tasks, similarity tasks, inference tasks, and reasoning tasks. 
To kick-start benchmarking for this language, these datasets were machine-translated from mainstream benchmarks in the literature and designated as \HyperGLUEpt.
We also make available baseline models for 10 of these tasks, developed with the low-rank adaptation approach over a state-of-the-art and open language model for Portuguese.

Even though MT datasets have their limitations and pitfalls, our manual analysis has found a relatively reduced amount of (translation and label) errors. We believe this renders our obtained datasets highly useful for assessing the comparative performance of neural language models for Portuguese.

In future work, it would be important to improve this benchmark with manual curation of the datasets (in particular, the test sets) and expand it with new ones.
Additionally, developing new datasets from scratch may better reflect the language and the cultures latent within language variants (which go well beyond European and Brazilian ones). Evolving these in a leaderboard would help foster research in the Portuguese language.

\section*{Acknowledgements}

This research was partially supported by:
PORTULAN CLARIN — Research Infrastructure for the Science and Technology of Language, funded by Lisboa 2020, Alentejo 2020 and FCT (PINFRA/22117/2016); ACCELERAT.AI - Multilingual Intelligent Contact Centers, funded by IAPMEI (C625734525-00462629);
ALBERTINA - Foundation Encoder Model for Portuguese and AI, funded by FCT (CPCA-IAC/AV/478394/2022);
Base Funding (UIDB/00027/2020) and Programmatic Funding (UIDP/00027/2020) of the Artificial Intelligence and Computer Science Laboratory (LIACC) funded by national funds through FCT/MCTES (PIDDAC);
Bernardo Leite is supported by a PhD studentship (with reference 2021.05432.BD), funded by Funda\c{c}\~{a}o para a Ci\^{e}ncia e a Tecnologia (FCT).

\bibliographystyle{apalike}  
\bibliography{main}

\end{document}